%% file: main.tex
\documentclass[letterpaper, 10 pt, conference]{ieeeconf}  

\let\labelindent\relax                                    
\IEEEoverridecommandlockouts                              

\input{preamble.tex}
\input{commands.tex}

\newcommand{\citet}[1]{\cite{#1}}
\newcommand{\citep}[1]{\cite{#1}}

\title{\LARGE \bf \titlelong}

\author{Cong Wang$^{1}$ \and Roberto Calandra$^{1}$ \and Verena Klös$^{2}$
\thanks{$^{1}$ LASR Lab,
          TU Dresden, Dresden, Germany
         {\tt\small \{cong.wang,roberto.calandra\}@tu-dresden.de}}%
\thanks{$^{2}$ Department of Computing Science,
          Carl von Ossietzky Universität Oldenburg, Oldenburg, Germany
        {\tt\small verena.kloes@uni-oldenburg.de}}%
}

\begin{document}

\maketitle
\thispagestyle{empty}
\pagestyle{empty}


\begin{abstract}
	\input{0_abstract.tex}
\end{abstract}


\section{INTRODUCTION}
\label{sec:introduction}

	\input{1_introduction.tex}


\section{RELATED WORK}
\label{sec:related}

	\input{2_related.tex}


\section{BACKGROUND: BDI-AGENT ARCHITECTURE}
\label{sec:background}

	\input{2b_background.tex}
	

\section{Study of User Preference: What is a good explanation?}  
\label{sec:study}

	\input{3a_study.tex}

        \input{3b_study_results.tex}

\section{Constructing effective explanations for BDI robots in HRI}
\label{sec:algorithm}
\input{4a_algorithm.tex}

 \section{IMPLEMENTATION \& EVALUATION}
 \label{sec:result}
 
 	\input{4b_result.tex}


 \section{CONCLUSION}
 \label{sec:conclusion}
 
 	\input{5_conclusion.tex}


\addtolength{\textheight}{-12cm}   





\section*{ACKNOWLEDGMENT}

\input{99_acknowledgments.tex}



\bibliographystyle{IEEEtran}
\bibliography{paper}

\end{document}

%% file: preamble.tex
\usepackage[english]{babel}

\usepackage{graphicx}
\usepackage{animate}
\usepackage{epsfig} 				
\usepackage{epstopdf}

\usepackage{listings}
\usepackage{color}
\usepackage{nameref}
\usepackage{hyperref}
\usepackage{amsmath}	 			
\usepackage{amssymb}  				

\usepackage{dsfont}			
\usepackage{mathtools}

\usepackage{epigraph}
\usepackage{lscape}
\usepackage[]{nomencl}				
\usepackage{algorithm}
\usepackage{algorithmic}
\usepackage{multicol}
\usepackage{multirow}
\usepackage{etoolbox}

\usepackage{caption}
\usepackage{subcaption}
\usepackage{wrapfig}

\usepackage{siunitx}

\usepackage{floatflt}

\usepackage{url}

\usepackage{enumitem}
\usepackage{subcaption}
\usepackage{booktabs}
\usepackage{cite}
\usepackage{comment}
\usepackage[dvipsnames]{xcolor}
    \definecolor{SoftAmber}{HTML}{886644}
    \definecolor{RoyalPurple}{HTML}{5B3B8C}
    \definecolor{SteelBlue}{HTML}{2A5CAA}
\usepackage[normalem]{ulem}
\usepackage{environ}

%% file: commands.tex
\newcommand{\email}[1]{\href{mailto:#1}{\nolinkurl{#1}}}

\newcommand{\alg}[1]{Algorithm~\ref{#1}}

\newcommand{\titlelong}[0]{Effective Explanations for Belief-Desire-Intention Robots:\\ When and What to Explain}

\newcommand{\todorv}[2]{\textcolor{red}{#1}\textsuperscript{\textcolor{orange}{(#2)}}}

\newcommand{\donerv}[3]{%
  #2%
}

\newif\ifshortenpaper
\shortenpapertrue

\NewEnviron{shortercomment}{
    \ifshortenpaper
    \else
        {\BODY}
    \fi%
}
\NewEnviron{longercomment}{
    \ifshortenpaper
        \BODY
    \fi
}

%% file: 0_abstract.tex
When robots perform complex and context-dependent tasks in our daily lives, deviations from expectations can confuse users. Explanations of the robot’s reasoning process can help users to understand the robot intentions.
However, when to provide explanations and what they contain are important to avoid user annoyance.
We have investigated user preferences for explanation
demand and content for a robot that helps with daily cleaning tasks in a kitchen. Our results show that users want explanations in surprising situations and prefer concise explanations that clearly state the intention behind the confusing action and the contextual factors that were relevant to this decision. 
Based on these findings, we propose two algorithms to identify surprising actions and to construct effective explanations for Belief-Desire-Intention (BDI) robots. Our algorithms can be easily integrated in the BDI reasoning process and pave the way for better human-robot interaction with context- and user-specific explanations.

%% file: 1_introduction.tex
Robots are becoming more present in our daily lives.
As robots perform more complex tasks and their behavior becomes context-dependent, robot behavior may deviate from human expectations and confuse users. Imagine the following scenario where a robot helps human users with their daily household tasks, such as putting away dishes or cleaning:

\emph{The user is in the kitchen with the robot and asks the robot to remove the used cups from the table and load them into the dishwasher. The robot is closer to the table where the cups are placed, so the user expects it to move directly to the table first.  Instead, the robot moves to the dishwasher and opens it, leaving the cups untouched on the table.}

In this situation, the robot does not behave as expected. This may not be a problem as long as the robot fulfills the task, but it can still be confusing for the user.
In such cases, an explanation of the robot's reasoning process can help to understand what the robot intends to do and to assess whether the behavior is correct and efficient\cite{anjomshoae2019explainable}. 
However, robots should not annoy humans with irrelevant explanations
\cite{wachowiak2024people}.
Thus, when to give an explanation and the content of the explanation have to be chosen carefully.

\begin{figure*}[t]
    \centering
    \includegraphics[width=0.9\linewidth,
                    trim={3.8cm 7.2cm 4.5cm 1.5cm},  
                    clip                     
                    ]{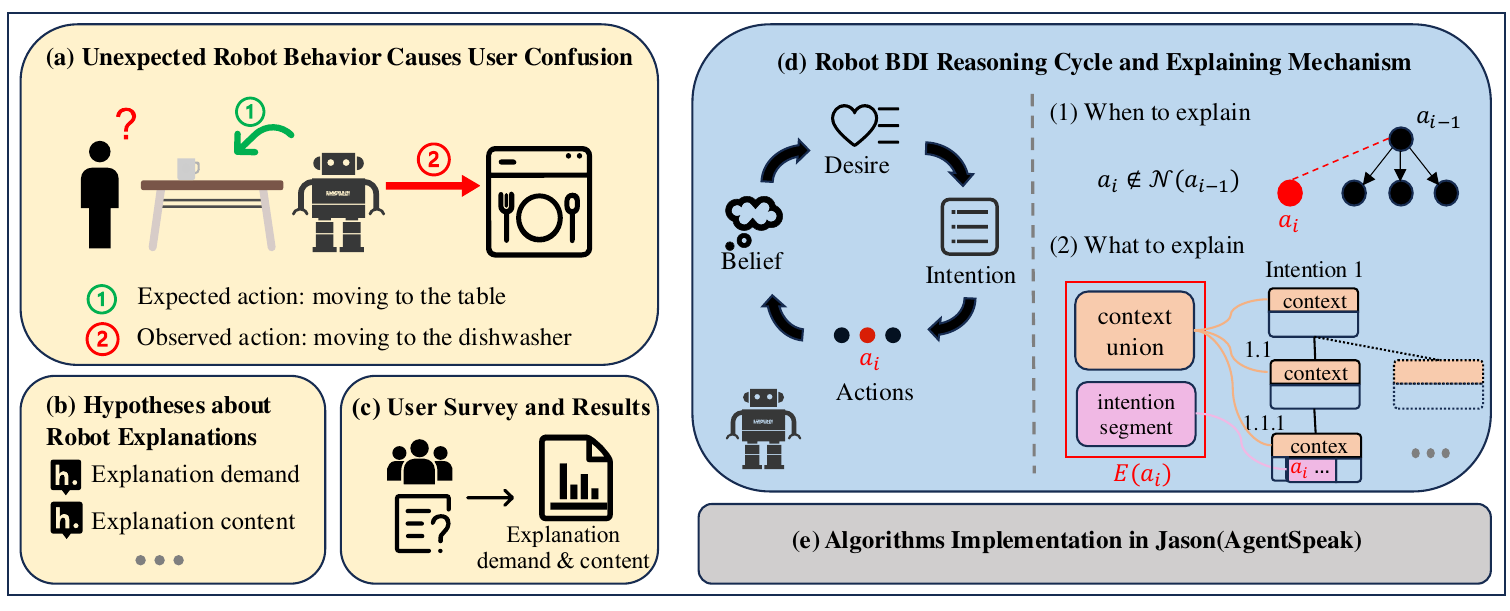}
    \caption{Our approach to explaining unexpected robot actions in a home-assistant scenario. (a) in our example, the robot’s behavior deviates from the user’s expectation after starting a cleaning task in a kitchen. (b) presents the main hypotheses about explanation demand and content. (c) represents the user survey phase. (d) highlights the BDI reasoning cycle with an explaining mechanism. The ``When to explain'' part detects unexpected actions, \textit{e.g.,} $a_i$ not in the expected successor group of last action $\mathcal{N}(a_{i-1})$. The ``What to explain'' part generates explanations $E(a_i)$ that contains key contextual factors from hierarchical reasoning. (e) we implement our algorithms in Jason (AgentSpeak).}
    \label{fig:figure1}
\end{figure*}

This paper addresses the open question of when 
robots should generate explanations and what content those explanations should include 
to align with user expectations in context-sensitive tasks (Fig.~\ref{fig:figure1}).
We conducted an online scenario-based user survey to investigate in which situations
users want an explanation from a home assistant robot in various kitchen cleaning tasks and what kind of explanation. 
To provide such explanations for surprising actions, 
we propose two algorithms that identify explanation demands and construct effective explanations for robots that are implemented as BDI agents. 
The construction identifies the relevant intention from a hierarchical goal structure and derives key contextual factors from hierarchical reasoning. 
We have implemented our algorithms in the BDI framework Jason\footnote{\url{https://jason-lang.github.io/}} to show its practical applicability~\cite{Jason}.
The full implementation is available online\footnote{\url{https://github.com/congw112358/RO-MAN_2025}}.
Our work lays the foundation for integrating user expectations into BDI-based robot reasoning, 
advancing explainable human–robot interaction (HRI) through context-aware explanations.

%% file: 2_related.tex
\textbf{Explaining Robot Behavior:}
Recent survey studies have identified explaining robot behavior as central in the growing field of explainable robotics. 
While the broader domain of explainable AI (XAI) has largely concentrated on making data-driven models interpretable, 
\cite{sado2023explainable} and \cite{sakai2022explainable} 
emphasize that explainable robots must communicate perceptual processes, 
cognitive reasoning, and behavioral decisions to human users.
From a social-cognitive perspective, the effectiveness of behavioral explanations depends on 
how well explanations align with human expectations. 
\cite{malle2006mind} and \cite{kashima1998category} argue that people interpret both robot and human behaviors via folk psychology,
including beliefs, desires, and intentions. 
\cite{miller2019explanation} consolidates findings from social science and psychology, 
highlighting that human explanations are causal, contrastive, and contextual. 
Recent work further demonstrates that people interpret robot behavior using folk-psychological constructs, 
in ways similar to their interpretations of humans
~\cite{thellman2022mental}.
These insights suggest that robotic explanations should not only expose internal processes, 
but also adhere to the explanatory norms familiar to humans.

\textbf{Mental-State-Based Explanation Mechanisms:}
Recent methods for robot explanations emphasize internal mental states (beliefs, goals, etc.) 
as the basis for generating cognitively aligned explanations.
\cite{sado2023explainable} survey techniques that expose agent beliefs and plans to users, 
while \cite{hanheide2017robot} integrate epistemic reasoning into planning, 
to explain assumptions and failures in uncertain environments. 
Although these methods improve structural transparency, they often overlook the interpretive gap between human and robot views.
Other work addresses this gap by modeling user mental states.
\cite{gao2020joint} propose hierarchical mind modeling for pedagogical explanations based on inferred user beliefs, 
and \cite{yuan2020joint} identify mismatches in belief attribution to generate target explanations.
However, few of these methods tackle the question of when to explain or what content best serves user needs during ongoing interactions.

\textbf{Explanation Demand and Content:}
Several studies explore explanation demand in human-robot collaboration. 
In a virtual search-and-rescue task, strategies such as explaining always, on request, or never
were compared, revealing that well-timed explanations improve team performance~\cite{chiou2022towards}. 
Similarly, user studies show that explanation demands vary across contexts. 
People expect explanations after errors or unexpected behavior, rather than routine success~\cite{wachowiak2024people}. 
Explanation content also matters.
\cite{sridharan2019towards} proposes a three-dimensional structure (decisions, underlying beliefs, and experiential basis) 
for multi-level explanations tailored to user needs.
Additionally, \cite{das2021explainable} find that contextual information and recent action history 
enhance user understanding, especially in failure recovery scenarios.
Nevertheless, we still lack adaptive mechanisms to determine both demand and content based on evolving user expectations.

\textbf{BDI Architectures for Robot Explanation:}
BDI architectures offer a structured foundation for explanation by modeling 
beliefs, desires, and intentions—concepts that align with human reasoning. 
\cite{harbers2010design} demonstrate that varying the abstraction level of goal representations 
affects user understanding, but their model does not account for explanation demand or context.
Trace-based approaches like \cite{koeman2019did} and \cite{dennis2022explaining} reconstruct reasoning processes or 
expose divergences in beliefs, but do not autonomously control when or what to explain.
\cite{ichida2023modeling} apply BDI modeling to task-oriented dialogue systems, 
enhancing response interpretability by linking utterances to underlying mental states.
\cite{winikoff2021bad} introduce “valuings” to express trade-offs in goal selection and evaluate them with a user study~\cite{winikoff2023evaluating}. 
Complementary work has explored multi-level frameworks to match explanation granularity to different user roles~\cite{yan2023towards}. 
While these efforts advance BDI-based explainability, they generally focus on design-time frameworks or post-hoc trace reconstructions,
leaving open how to dynamically manage explanation demand and content in real-world HRI.

Prior work has established the importance of goal-driven, cognitively aligned robot explanations. 
However, existing approaches provide limited support for determining 
in which situations users prefer explanations or for selecting relevant content based on their expectations. 
This gap motivates our approach to construct demand-driven and effective explanations for BDI agents in interactive HRI settings.

%% file: 2b_background.tex
The BDI architecture is a widely used framework for designing intelligent agents. 
It models behavior based on the agent's \emph{Beliefs} about the world,
\emph{Desires} representing its goals and \emph{Intentions}, 
the plans it commits to in order to achieve those goals.
There exist several implementations of this architecture. We use the multi-agent system development platform JASON 
that provides an interpreter for an extension of the BDI-agent programming language AgentSpeak \cite{AgentSpeak}.
We now provide a formal model of a BDI robot that is based on AgentSpeak. 
Our formal model is indicated by
$M =\langle S, B, G,\Pi\rangle$,
where $S$ is the set of all possible states, $B$ the robot’s current beliefs, 
$G$ its goals, and $\Pi$ a library of plan templates. 

\emph{State Space:}
We let
$
S \;=\;\bigl(S_{\mathrm{env}} \times S_{\mathrm{obj}} \times S_{\mathrm{robot}} \times S_{\mathrm{user}}\bigr)
\;=\;\prod_{v\in\mathcal{V}}\mathrm{Dom}(v),
$
where the variables are conceptually partitioned into four groups: 
$\mathrm{env}$ (environment), $\mathrm{obj}$ (object properties), 
$\mathrm{robot}$ (robot internals) and $\mathrm{user}$ (user-related data). 
We employ a first-order logic language $\mathcal{L}(S)$ over $S$ to describe facts or conditions. 
For a formula $\varphi \in \mathcal{L}(S)$, writing $s \models \varphi$ means that $\varphi$ is true in state $s$.

\emph{Beliefs and Goals:} 
\emph{Beliefs} ($B$) represents the robot’s internal knowledge or assumptions, 
while \emph{Goals} ($G$) captures the intended results of the robot. 
We do not distinguish between \textit{Desires} and \textit{Goals} in this work
because our scenarios involve a fixed set of tasks where all desires are immediately adopted as goals.
This simplification is frequently adopted in BDI agent applications, 
allowing us to focus on the explanation mechanisms.
Concretely, let
$
B \subseteq \mathcal{L}(S),
$
and
$
G \subseteq \mathcal{L}(S).
$

\emph{Events:}
A set of events $\mathcal{E}$, similar to events defined in the AgentSpeak, 
stands for symbolic triggers for adding or removing $B$ or $D$.
These events serve as triggers that prompt the robot to search for a matching plan in the plan library.

\emph{Plan Library: }
The set $\Pi$ is a \emph{plan library}, comprising plan templates.
Each plan $\pi \in \Pi$ is defined as
$
\pi =(\tau,\;\gamma,\;\beta),
$
and we refer to $\tau(\pi)$, $\gamma(\pi)$, $\beta(\pi)$ for clarity. Here:
$\tau \in \mathcal{E}$ is the \textit{trigger event},
$\gamma \subseteq \mathcal{L}(S)$ is the \textit{context condition}, 
$\beta$ is the \textit{body}, a sequence of steps. 
Each step is either an action or a sub-intention. 

\emph{Intention:}
Once an occurring event $\varepsilon \in \mathcal{E}$ matches the trigger $\tau(\pi)$ 
of some plan $\pi \in \Pi$, and if $B \models \gamma(\pi)$, 
the robot \emph{adopts} $\pi$ as its intention. 
Whenever no explicit substitution is required, we employ the simplified notation
$
I=\pi.
$
If there exists some intention $\pi_j$ such that $\pi_i \in \beta(\pi_j)$, 
then $\pi_j$ is called the \emph{direct parent} of the sub-intention $\pi_i$. 
If no such $\pi_j$ exists, we call $\pi_i$ a \emph{root} intention.

\emph{Action:}
Each action $a$ in the action set $\mathcal{A}$ has
preconditions $\mathrm{Pre}(a)\subseteq\mathcal{L}(S)$
and an effect
$\mathrm{Eff}^\pm(a)\subseteq\mathcal{L}(S)$.

%% file: 3a_study.tex
\textbf{Exploratory Hypotheses:}
To improve human-robot interaction, it may be helpful to provide an explanation for the robot behavior. However, not all explanations may be perceived as helpful. 
In this exploratory study, we examine in which situations explanations are desired and what types of explanations are perceived as most effective.
We examine the following exploratory hypotheses concerning how users perceive explanations for robot behavior:

\begin{enumerate}[label=(H\arabic*), ref=H\arabic*]
    \item\label{hyp:one} Higher surprise is associated with a greater likelihood that a user wants an explanation.
    \item\label{hyp:two} Explanations that include contextual factors are perceived as more useful than all other explanation types.
    \item\label{hyp:three} Explanation contents are more effective when they clearly identify the key contextual factors
    (\textit{i.e.,} the essential beliefs or context conditions, that underlie the robot’s current plan).
    \item\label{hyp:four} Explanations should focus on the portion of the plan that is directly relevant to the key contextual factors, rather than detailing redundant actions.
\end{enumerate}

While ~\ref{hyp:one} concerns the demand for an explanation, 
\textit{i.e., when} the robot should provide an explanation to a human, 
~\ref{hyp:two}, \ref{hyp:three} and \ref{hyp:four} address the content, 
\textit{i.e., what} constitutes an effective explanation.

\begin{longercomment}
\begin{figure}[t]
    \centering
    \setlength{\fboxrule}{0.9pt} 
    \setlength{\fboxsep}{1pt} 
    \fbox{
        \includegraphics[
        width=0.8\linewidth,
        trim={2.0cm 7.0cm 2.0cm 11.2cm},  
        clip                     
        ]{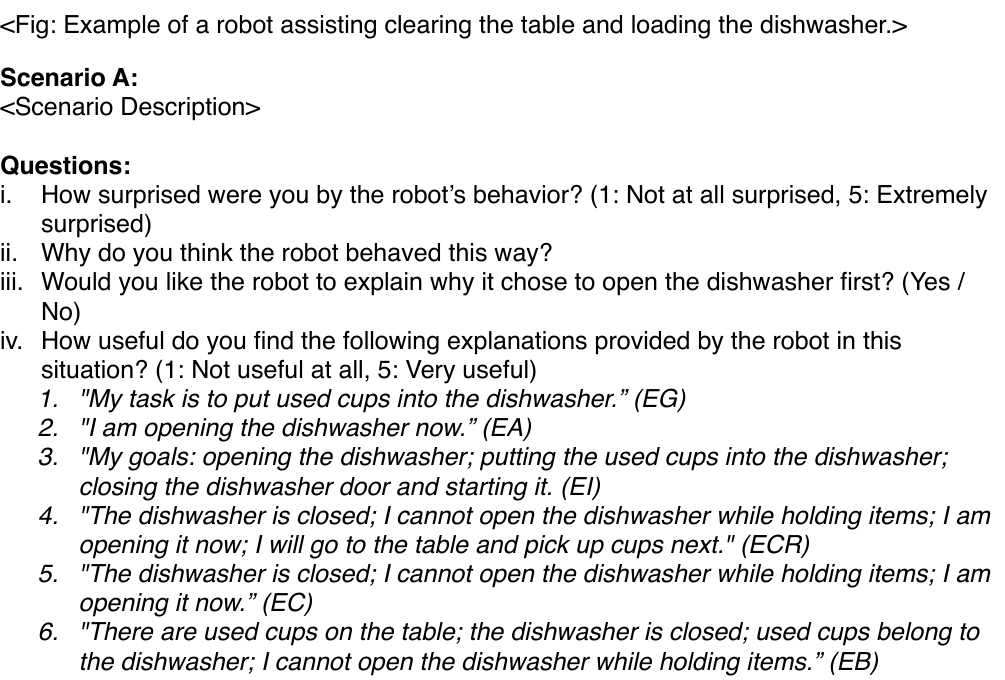}
    }
    \caption{Illustrative questions from the questionnaire.} 
    \label{fig:questionnaire-no}
\end{figure}

\end{longercomment}

\begin{shortercomment}
\begin{figure}[t]
    \centering
    \setlength{\fboxrule}{0.8pt} 
    \setlength{\fboxsep}{1pt} 
    \fbox{
        \includegraphics[
        width=0.82\linewidth,
        trim={2cm 6.8cm 2cm 2.7cm},  
        clip                     
        ]{ROMAN2025/figs/Questionnaire_Sample_new_NoDescription.pdf}
    }
    \caption{An example of one scene in our questionnaire. The robot (Pepper) is positioned beside a table in a typical kitchen layout, thereby illustrating a scenario for testing participants’ reactions to the robot’s unexpected behavior.}
    \label{fig:questionnaire-no}
\end{figure}
\end{shortercomment}

\textbf{Study Design:}

\textit{Participants:}
Thirty-three participants ($N = 33$) from multiple academic institutions, 
including students, doctoral researchers, and faculty members,
took part in the study.
Participants were aged 18--64 years 
(estimated $M = 34.4$ years, $SD = 9.8$ based on bin midpoints); 70\% were aged 25--34.
All participants were volunteers and encompassed a range of academic disciplines.
Participants received a short overview of the study and accessed the online questionnaire via a link, 
indicating consent by continuing.

\textit{Materials:}
We adopted a scenario-based approach to examine participants’ responses to robot behaviors and explanations. 
A static image was used to depict the kitchen environment and the assistant robot.
Each of the six scenarios was then introduced through a unique textual description 
outlining a specific household task
along with a likely unexpected robot action.
Participants responded to a subset of six scenarios, either selecting 
how many to answer based on their available time (with a minimum of three), 
or completing three randomly assigned scenarios.
All scenarios were presented in randomized order, and only one was visible at a time.
The questionnaire was deployed as an online survey, which supported random selection of scenarios, 
randomized scenario presentation and controlled scenario visibility. 
For each unexpected action, participants rated the usefulness of six possible explanations, 
which contain the unexpected action (EA), the high-level goal (EG), 
key contextual factors (EC), key contextual factors and a redundant action (ECR), current beliefs (EB), or current intentions (EI). 
Abbreviations of the explanation types are listed in Table~\ref{tab:what}. 
An example item from the questionnaire is shown in Fig.~\ref{fig:questionnaire-no}; 
the full questionnaire is available in our online repository.

\textit{Variables:}
We measured the following variables in each scenario:
    a) Perceived surprise: Participants rated how surprised they were by the robot’s behavior on a 5-point Likert scale (1 = Not at all surprised, 5 = Extremely surprised). This variable was used as a predictor in testing \ref{hyp:one}.
    b) Desire for explanation: After observing the robot's behavior, participants were asked whether they would like the robot to explain its action (Yes/No). This binary response was used as the outcome variable in \ref{hyp:one}.
    c) Usefulness ratings: For each scenario, participants were presented with six explanations corresponding to different explanation types. They rated the usefulness of each explanation on a 5-point Likert scale (1 = Not useful at all, 5 = Very useful). These ratings were analyzed in relation to Hypotheses \ref{hyp:two}, \ref{hyp:three} and \ref{hyp:four}.

\textit{Design:}
We designed a user survey to investigate explanation demand and preferred content in household HRI scenarios.
The survey focused on participants' responses to unexpected robot behavior and on 
what kind of explanation contents are most useful.

\textit{Procedure:}
Participants began with a brief introduction and a set of background questions, 
then proceeded to the main task involving multiple scenarios. 
They answered all questions within each scenario before proceeding to the next.

%% file: 3b_study_results.tex
\begin{table}[t]
\centering
\caption{Logistic Regression Results: Relationship between Explanation Demand and Surprise}
\label{tab:when}
\begin{tabular}{lccc}
\toprule
\textbf{Variable} & \textbf{OR (95\% CI)} & \textbf{\emph{p}-value} \\ \hline
\midrule
Intercept         & 0.22 (0.07, 0.70)     & 0.011                                    \\
Surprise          & 3.30 (1.93, 5.62)     & \textless{}0.001                         \\ \hline
\bottomrule
\end{tabular}

\vspace{1ex}
\footnotesize
\textbf{Notes:} 
Sample size \(N=120\). Pseudo \(R^2 = 0.2145\). LLR \(p\)-value = \(1.61\times 10^{-7}\). 
Intercept \(\approx 0.22\) and Surprise \(\approx 3.30\).
\end{table}

\textbf{Study Results:}
Building on~\ref{hyp:one} that higher surprise trigger a stronger demand for explanations,
we conducted a logistic regression to examine whether participants' 
surprise regarding the robot's 
behavior (rated on a 5-point scale) predicts 
their demand for explanations 
(binary outcome: 0 = no, 1 = yes).
The results indicate that higher levels of surprise were significantly associated 
with a greater likelihood of requesting an explanation (\(p < 0.001\)).
A one-point increase in the surprise rating yielded an odds ratio (OR) 
of 3.30, 95\% CI [1.93, 5.62], suggesting that participants were more than three times 
as likely to desire an explanation when they felt one unit higher on the surprise scale.
The model was statistically significant (LLR \(p\)-value = \(1.61 \times 10^{-7}\)) 
with a pseudo \(R^2\) of 0.2145, indicating that the reported surprise accounts 
for a notable portion of the variance in the demand for explanations.%
%
\begin{table}[t]
\centering
\caption{Usefulness Scores of Different Explanation Types with Descriptions}
\begin{tabular}{cccc}
\toprule
\textbf{Type} &\textbf{Abbreviation Description} & \textbf{Mean} & \textbf{Std} \\ \hline
\midrule
EA         & contains a specific action               & 1.530         & 0.882        \\
EG         & contains a high-level goal               & 1.591         & 0.887        \\
EC         & contains key contextual factors          & 4.383         & 0.874        \\
ECR        & EC plus a following redundant action     & 4.061         & 1.011        \\
EB         & contains extensive current beliefs       & 3.878         & 1.085        \\
EI         & contains extensive current intentions    & 3.339         & 1.199        \\ \hline
\bottomrule
\end{tabular}
\label{tab:what}
\end{table}
%
\begin{figure}[t]
    \centering
    \includegraphics[width=0.8\linewidth]{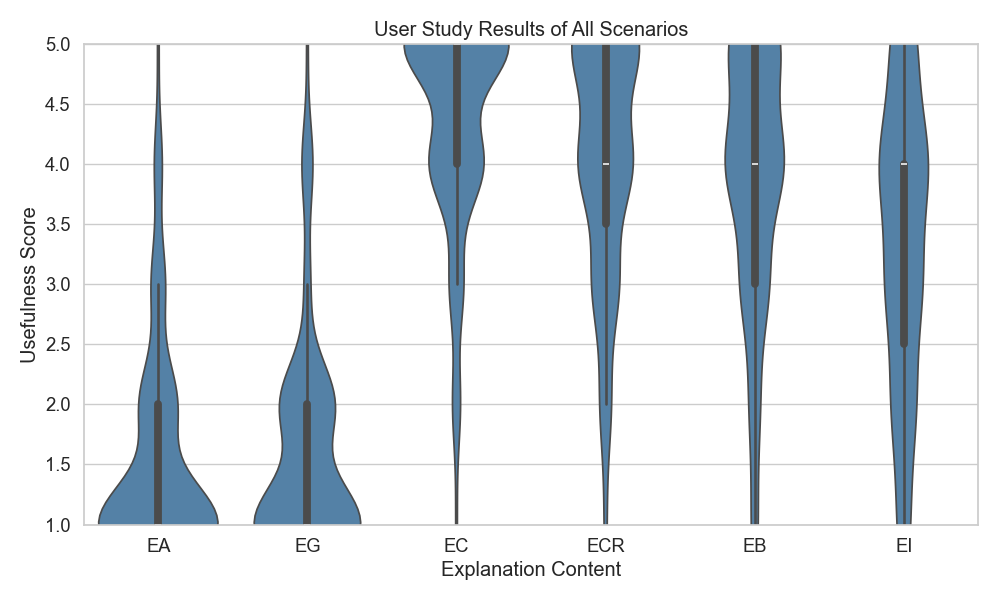}
    \caption{Distribution of usefulness scores for six explanation content types in the user survey. Violin plots show ratings distribution. EC and ECR scored highest, with EC slightly ahead. EA and EG received consistently lower ratings.}
    \label{fig:unified-violin}
\end{figure}
Table~\ref{tab:what} presents the mean and standard deviation of usefulness ratings (1–5 scale)
for six different explanation types.
Figure~\ref{fig:unified-violin} shows the corresponding violin plots,
illustrating both the central tendency and the distribution of responses.

\textbf{Discussion:}
Both EA 
and EG 
received low mean ratings, 1.53 and 1.59 respectively. 
This suggests that explanations narrowly focused on a single action (EA) or on a 
high-level goal (EG) do not sufficiently address participants’ informational needs. 
EC 
attained the highest mean rating (4.38) 
with a relatively small standard deviation (0.87), 
indicating that participants overwhelmingly perceived it as useful. 
In Fig.~\ref{fig:unified-violin}, EC’s distribution skews toward the upper end of the scale, 
suggesting broad agreement among participants.
ECR 
averaged 4.06. 
Although it remains relatively high, the additional 
non-essential or redundant action 
may have modestly reduced its clarity or perceived usefulness compared to EC.
Both EB (3.88) and EI (3.34) show lower usefulness scores than EC. 
While they provide extensive information about the robot’s beliefs or intentions, 
they do not isolate the specific contextual factors most pertinent to the current action. 
Explanations narrowly limited to a single action (EA) or framed around a broad, general goal (EG) were found to be less effective, 
which aligns with~\ref{hyp:two}. 
In contrast, highlighting the essential contextual factors (EC) emerged as most beneficial, confirming~\ref{hyp:three}. 
Including extraneous details such as redundant actions (ECR) or extensive information 
about beliefs and intentions (EB, EI) diluted clarity, further supporting~\ref{hyp:four}. 
Overall, these results underscore that emphasizing the key contextual factors is crucial in robotic explanations 
to better address human’s informational needs.

%% file: 4a_algorithm.tex
From our study in Section \ref{sec:study}, we learned that home assistant robots should give explanations in unexpected/surprising situations and that these explanations should contain the key contextual factors regarding the current beliefs and the associated intentions.
In the following,
we present our algorithms to decide which actions require an explanation (cf. Sec.~\ref{ssec:when}) 
and to construct effective explanations for those actions (cf. Sec.~\ref{ssec:what}).
Our goal is twofold: 
to identify \textit{when} the robot should provide an explanation for its behavior (i.e., which actions appear surprising to the user) 
and \textit{what} the robot should include in its explanation (i.e., which beliefs, intentions, and contextual factors are most relevant to convey). 
Our mechanism integrates two main components:
\begin{itemize}
    \item Explanation Demand: We propose a lightweight user-modeling approach (detailed in Sec.~\ref{ssec:when}) to detect when an action may appear \emph{unexpected} or \emph{surprising} from the user’s viewpoint. This triggers an explanation to clarify the robot’s current or upcoming behavior.
    \item Explanation Content: Once an explanation is triggered, the system automatically collates (i) the relevant context conditions inherited from the hierarchy of active intentions and (ii) the subsequent actions aimed at altering these conditions, as elaborated in Sec.~\ref{ssec:what}. By referencing the agent’s key contextual factors, we ensure that explanation content captures both \emph{what} the robot is doing and \emph{why}.
\end{itemize}

\subsubsection{Explanation Demand}\label{ssec:when}
To decide whether an action should be explained or not, we have to decide whether it is surprising to the human. We assume that humans expect certain actions either because they are directly linked to the order that the human gave to the robot (e.g., for the order to clean the kitchen, the robot is expected to move to the kitchen if not already there), because two actions frequently occur together (e.g., opening a door and moving through the door), or due to previously observed action sequences.
To model that, we define a function
$
\mathcal{N} : \mathcal{A} \;\to\; 2^{\mathcal{A}}
$
that maps each action $a$ to a set $\mathcal{N}(a)$ of \textit{expected successors}. 
At runtime, if the next action is not in $\mathcal{N}(a)$, 
we trigger an explanation and add the action to that set (to learn from observed behavior).
If the agent starts with an empty set of \textit{expected successors} for each action, all actions will be explained at their first occurrence. However, $\mathcal{N}$ can also be initialized with actions that frequently occur together
or with expected pairs that belong to the current task. This explicit modeling of \textit{expected successors} is a simple model of the human's mental model and can also be used to differentiate between different users by having specific functions for each user. Thus, capturing individual experiences or context knowledge, e.g., user A gave the order to clean the kitchen but user B does not know why the robot moves to the kitchen.  

At runtime, whenever a robot adopts a new plan as intention, we use \alg{alg:when} to mark all actions in the plan that require an explanation, and before such an action 
is executed, \alg{alg:what} is used to create the explanation.
In \alg{alg:when}, let the current action $a_i$ be part of the intention $\pi_i$ 
with body $\beta(\pi_i)=(a_1,a_2,\dots,a_n)$. 
For each pair of consecutive actions $(a_i,\,a_{i+1})$, 
the robot checks whether $a_i\in\mathcal{N}(a_{i+1})$ (line 3) and if not, triggers an explanation (line 4) and adds the action to $\mathcal{N}$ (line 5).

\begin{algorithm}[t]
\caption{Explanation Demand}
\label{alg:when}
\begin{algorithmic}[1]
\REQUIRE Action $a_i$ in the body $\beta(\pi_i)=(a_1,\dots,a_n)$,
         expected successor $\mathcal{N}:\mathcal{A}\to 2^{\mathcal{A}}$.
\ENSURE Provide an explanation if the next action is not in $\mathcal{N}(a_i)$.

\FOR{$i \gets 1$ to $n-1$}
   \STATE $\mathrm{succ} \gets a_{i+1}$
   \IF{$\mathrm{succ}\,\notin\,\mathcal{N}(a_i)$}
      \STATE \textbf{explanation triggered:}
             // e.g. ``$a_{i+1}$ is unexpected after $a_i$!''
      \STATE $\mathcal{N}(a_i) \gets \mathcal{N}(a_i)\cup\{\mathrm{succ}\}$
   \ENDIF
\ENDFOR

\end{algorithmic}
\end{algorithm}

\subsubsection{Explanation Content}\label{ssec:what}

Useful explanations should contain the key contextual factors
that led to the confusing action of the robot. 
In our example (cf. Sec.~\ref{sec:introduction}), the confusing behavior is \texttt{navigateTo(dishwasher)} and an explanation should contain the beliefs about the context that triggered this behavior (\texttt{dishwasherDoor(closed)}) and the actual intention of moving to the dishwasher, which is the action of the current plan that modifies this key belief (\texttt{openDoor(dishwasher)}). 
As intentions can be hierarchical, 
(e.g., \texttt{openDishwasherIfNeeded}  as subintention of \texttt{storeUsedCup}),
context conditions of parent intentions are also important to explain the adoption of the current intention. 
We define the set of \emph{all} parent intentions of an intention $\pi_i$, 
denoted by $\mathcal{P}(\pi_i)$, as follows:
\[
\mathcal{P}(\pi_i) = 
\begin{cases}
\varnothing, & \text{if $\pi_i$ is a root intention,}\\
\{\pi_j\}\,\cup\,\mathcal{P}(\pi_j), & \text{if $\pi_j$ is the direct parent of $\pi_i$.}
\end{cases}
\]

Let $\pi_i$ be the current intention and 
$\mathcal{P}(\pi_i)$ 
the parents of $\pi_i$. We define
$\Gamma_i \;=\; \bigcup_{\pi \in \mathcal{P}(\pi_i)} \gamma(\pi),$
which is the \emph{hierarchical intention context}, gathering all context conditions from each intention in the parent intention chain.
\alg{alg:what} explains the triggered action $a_i$, 
by identifying the key hierarchical intention context and the associated intention.
It first collects the hierarchical intention context from the action's intention and the associated parent intentions (line 1).
Then, it identifies the next action in the set of steps that modifies these hierarchical context conditions (lines 2-9) and returns the intention context together with the sequence of actions that will be performed to change the context condition (line 10).
In doing so, the algorithm integrates high-level context with the local action sequence.

\begin{algorithm}[t]
\caption{Explanation Content}
\label{alg:what}
\begin{algorithmic}[1]
\REQUIRE An action $a_i$ in the body $\beta(\pi_i)=(a_1,\dots,a_n)$ of the current intention $\pi_i$. 
         The $\mathcal{P}(\pi_i)$. 
\ENSURE $\mathrm{E}(a_i) = \Bigl(\Gamma_i,\;(a_i,a_{i+1},\dots,a_k)\Bigr)$.

\STATE $\displaystyle \Gamma_i \gets \bigcup_{\pi \in \mathcal{P}(\pi_i)} \gamma(\pi)$ 
   \quad // hierarchical union of key context conditions
\STATE $k \gets n$

\FOR{$t \gets i$ \TO $n$}
   \STATE \textit{// check if $a_t$ modifies $\Gamma_i$}
   \IF{$\bigl(\mathrm{Eff}^+(a_t)\cup \mathrm{Eff}^-(a_t)\bigr)\;\cap\;\Gamma_i \;\neq\;\varnothing$}
      \STATE $k \gets t $
      \STATE \textbf{break}
   \ENDIF
\ENDFOR

\RETURN $\Bigl(\Gamma_i,\;(a_i,a_{i+1},\dots,a_k)\Bigr)$

\end{algorithmic}
\end{algorithm}

\begin{shortercomment}
To illustrate our algorithm, we apply it on our running example. 
Suppose the robot's current \emph{Belief} $B$ is:
\{
$\mathsf{dishwasherDoor}(\mathsf{closed})$,
$\mathsf{cup}_{1}(\mathsf{used}, \mathsf{table})$,
$\mathsf{cup}_{2}(\mathsf{clean}, \mathsf{table})$,
$\mathsf{robot} (\mathsf{table}, \mathsf{holding}(\mathsf{none}))$\}.
The robot's \emph{goal} $G$ is to place the used cup into the dishwasher: 
$G = \{\mathsf{cup}{1}(\mathsf{used}, \mathsf{dishwasher})\}$,
The robot chooses \emph{plan $\pi_1$} with sub-intention \emph{plan $\pi_2$} to achieve the goal. \\
{\small
\emph{Plan}\;\(\pi_{1}\)\textbf{:}\;\texttt{storeUsedCup.}
\begin{align*}
\tau(\pi_1) \;=\;& +!\mathsf{storeCup}(C), \\
\gamma(\pi_1) \;=\;& \{\mathsf{used}(C))\}, \\
\beta(\pi_1) \;=\;& 
    \begin{aligned}[t]
        \Bigl(
       &+!\mathsf{openDishwasherIfNeed},\; \mathsf{navigateTo}(\mathsf{table}), \\
       &\mathsf{pickUp}(C),\; \mathsf{navigateTo}(\mathsf{dishwasher}), \mathsf{putDown}(C)
        \Bigr).
     \end{aligned}
\end{align*}
\emph{Plan}\;\(\pi_{2}\)\textbf{:}\;\texttt{openDishwasherIfNeed.}
\[
\begin{aligned}
\tau(\pi_2) \;=\;& +!\mathsf{openDishwasherIfNeed}, \\
\gamma(\pi_2) \;=\;& \{\,\neg\mathsf{dishwasherDoor}(\mathsf{open}), \mathsf{holding}(\mathsf{none})\}, \\
\beta(\pi_2) \;=\;& 
   \bigl(
     \mathsf{navigateTo}(\mathsf{dishwasher}),\;
     \mathsf{openDoor}(\mathsf{dishwasher})
   \bigr).
\end{aligned}
\]
}
To construct an explanation with \alg{alg:what}, we first construct parents $\mathcal{P}(\pi_2) = \{\pi_2, \pi_1\}$ and the hierarchical intention context 
$\Gamma_i= \{\mathsf{used}(C), \mathsf{dishwasherDoor}(\mathsf{closed}), \mathsf{holding}(\mathsf{none}) \}$. 
Since 
$
\mathrm{Eff}^\pm(\mathsf{openDoor}(\mathsf{dishwasher})) \cap\;\Gamma_i \;\neq\;\varnothing
$, with 
$\mathrm{Eff}^\pm(\mathsf{openDoor}(\mathsf{dishwasher}))= \{\,\mathsf{dishwasher}(\mathsf{open})$, $\mathsf{dishwasher}(\mathsf{closed})\}$, 
we get $a_t = \text{openDoor}(\mathsf{dishwasher}) $ and explanation content 
\begin{multline*}
E(\mathsf{navigateTo}(\mathsf{dishwasher})) =  \\
\{(\mathsf{used}(C), \mathsf{dishwasherDoor}(\mathsf{closed}), \mathsf{holding}(\mathsf{none})), \\
(\mathsf{navigateTo}(\mathsf{dishwasher}), \mathsf{openDoor}(\mathsf{dishwasher}))  \}
\end{multline*}
\end{shortercomment}

%% file: 4b_result.tex
To demonstrate 
the practical effectiveness of our explanation framework,
we developed a prototype system using JASON~\cite{Jason}, a widely used BDI-agent programming platform. 
This system simulates a household robot scenario, mirroring the tasks and environments introduced in our user study.
To embed Algorithms~\ref{alg:when} and \ref{alg:what} in a BDI framework, we utilize a predefined belief list that represents expected behaviors. 
When an upcoming event is not in the list, it is considered unexpected.
Afterwards, the agent spawns a sub-intention for explanation 
by collecting the hierarchical context conditions and future relevant actions in the current intention.
Our prototype demonstrates how the surprise-driven explanation trigger and key-context-factors-focused explanation content 
can be seamlessly integrated into a BDI agent. 
By leveraging JASON’s event-based structure and plan hierarchy, 
we efficiently detect unexpected actions and generate human-readable justifications.
The full implementation is available online in our repository.
Overall, these solutions confirm that the agent’s explanations matches the design goals derived from our earlier survey (Section~\ref{sec:study}), 
and they illustrate the practical feasibility of operationalizing Algorithm~\ref{alg:when} 
and~\ref{alg:what} within a BDI framework. 
The prototype effectively implements our explanation methodology in a straightforward manner
and can be extended in future work to a real household robot scenario.

%% file: 5_conclusion.tex
We have investigated
in which situations users want an explanation and what content they prefer,
across several scenarios where a robot helps humans with daily cleaning tasks in a kitchen. 
Results show 1) a positive correlation between surprise and the demand for an explanation and 2) that users prefer concise explanations that clearly state the intention behind the confusing action and the contextual factors that were relevant for this decision. 
Based on these findings, we proposed two algorithms to identify surprising actions and to construct effective explanations by identifying the relevant intention from a hierarchical intension structure and deriving key contextual factors from hierarchical reasoning of robots that are implemented as BDI agents. We implemented our algorithms in the BDI framework JASON to show its practical applicability. 
Future directions include: 
explore the timing and necessity of explanations in different interaction contexts, evaluate the proposed explanation methods through experiments with 
real-world robotic platforms and a broader range of users, and investigate user-specific explanations.

%% file: 99_acknowledgments.tex
This work is partly supported by the German Research Foundation (DFG 
) as part of Germany’s Excellence Strategy – EXC 2050/1 – Project ID 390696704 – Cluster of Excellence “Centre for Tactile Internet with Human-in-the-Loop” (CeTI)
of Technische Universität Dresden,
by the project ``Genius Robot'' (01IS24083) funded by the Federal Ministry of Education and Research (BMBF), and by the German Academic Exchange Service (DAAD) in project 57616814 (\href{https://secai.org/}{SECAI} \href{https://secai.org/}{School of Embedded and Composite AI}).